\documentclass{article} 
\usepackage{iclr2027_conference,times}

\usepackage{amsmath,amsfonts,bm}

\def\eqref#1{equation~\ref{#1}}

\def\1{\bm{1}}

\DeclareMathAlphabet{\mathsfit}{\encodingdefault}{\sfdefault}{m}{sl}
\SetMathAlphabet{\mathsfit}{bold}{\encodingdefault}{\sfdefault}{bx}{n}

\usepackage{caption}
\usepackage{url}
\usepackage{booktabs}
\usepackage{multirow}
\usepackage{graphicx}
\usepackage{amsmath}
\usepackage{amssymb}
\usepackage{algorithm}
\usepackage{algorithmic}
\usepackage{colortbl}
\usepackage{tabularx}
\newcolumntype{Y}{>{\centering\arraybackslash}X}

\usepackage{xcolor}

\definecolor{lightblue}{RGB}{83,136,233}
\definecolor{darkblue}{RGB}{32,49,154}
\colorlet{modelcolor}{darkblue}

\usepackage[colorlinks, linkcolor=darkblue, urlcolor=darkblue,citecolor=darkblue]{hyperref}

\newcommand{\method}{L2R}
\newcommand{\methodshort}{L2R}

\title{Learning to Retrieve: Internalizing Memory Retrieval for Video World Models}

\author{
JiaKui Hu$^{1,2}$\thanks{Equal contribution.}\hspace{4mm}
Tailai Chen$^{3 \ *}$\hspace{4mm}
Yuqi Pan$^{3}$\hspace{4mm}
Xuerui Qiu$^{3}$\hspace{4mm}
Jialun Liu$^{4}$\hspace{4mm}
Xiao Cao$^{5}$\hspace{4mm}
\\\
\textbf{Zhenxin Zhu}$^{2}$\hspace{4mm}
\textbf{Guang Chen}$^{2}$\hspace{4mm}
\textbf{Hangjun Ye}$^{2}$\hspace{4mm}
\textbf{Bing Wang}$^{2}$\thanks{Project leader.}\hspace{4mm}
\textbf{Yanye Lu}$^{1}$\thanks{Corresponding author.}
\\
$^{1}$\normalfont{Peking University}\hspace{4mm}
$^{2}$\normalfont{Xiaomi EV}\hspace{4mm}
$^{3}$\normalfont{CASIA}\hspace{4mm}
$^{4}$\normalfont{UQ}\hspace{4mm} 
$^{5}$\normalfont{NUS}\hspace{4mm} 
\\
}

\iclrfinalcopy 

\begin{document}


\maketitle
\vspace{-3.3em}
\begin{center}
    \url{https://jkhu29.github.io/l2r}
\end{center}

\vspace{-4mm}
\begin{center}
\includegraphics[width=\textwidth]{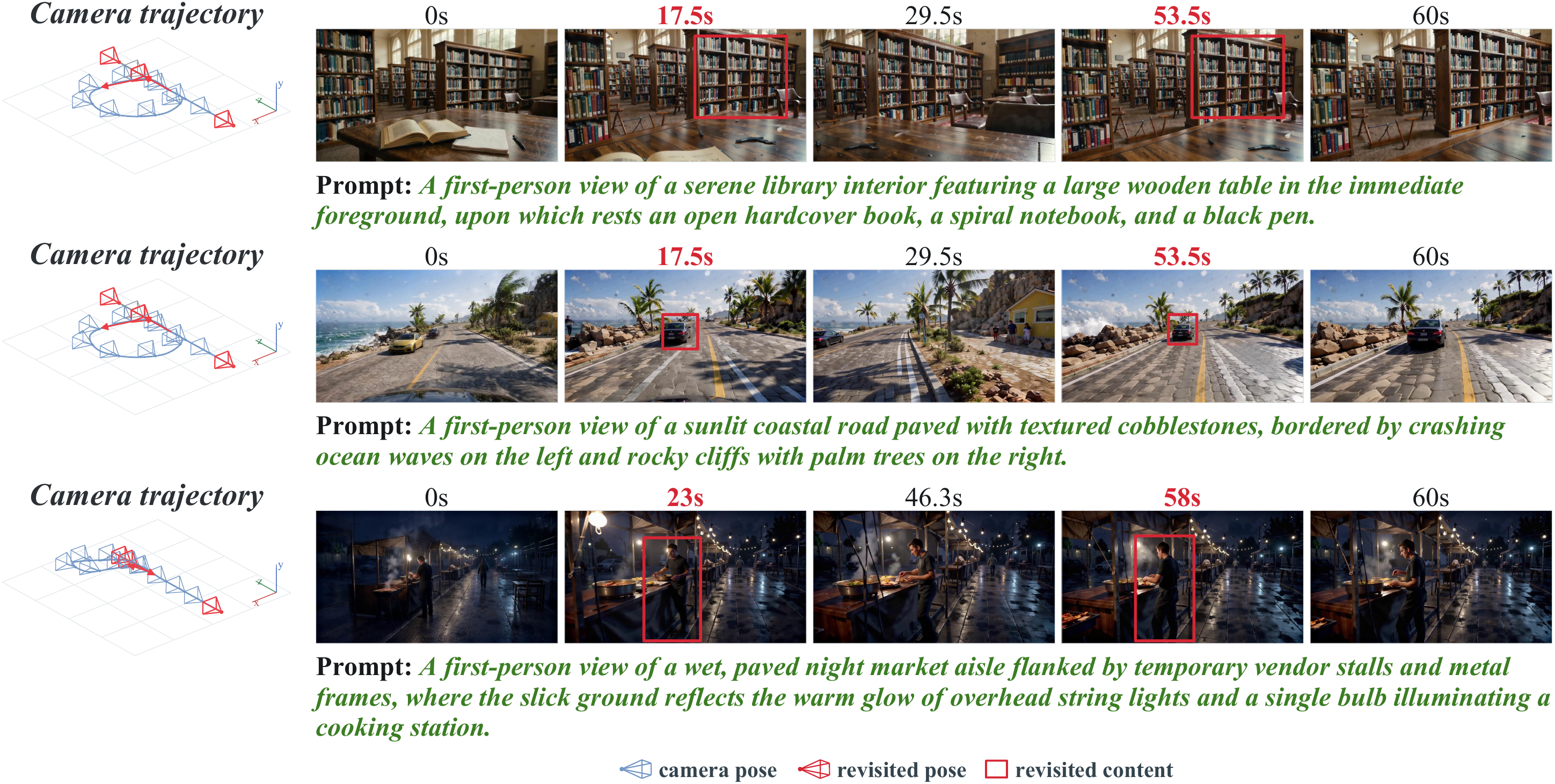}
\captionof{figure}{
\textbf{Teaser demonstration.} We introduce Learning-to-Retrieve (L2R), which can generate consistent videos without external memory modules or 3D reconstruction pipelines. Red boxes mark content observed from matched camera poses separated by a long temporal interval, illustrating that \method{} preserves scene identity when the camera revisits a previously observed region.
}
\label{fig:teaser}
\end{center}

\begin{abstract}
Video world models aim to generate explorable, 3D-consistent scene videos conditioned on camera trajectories. Existing approaches often rely on external memory systems that explicitly retrieve previously observed content to mitigate scene drift during long-horizon generation. However, these auxiliary memory pathways operate outside the model's internal generative dynamics, preventing the model from intrinsically learning when and what historical information should be retrieved. We propose to internalize memory retrieval into the generation process, allowing retrieval to emerge as an intrinsic behavior of the video world model rather than relying on an external memory system. Based on this principle, we introduce \textbf{Learning-to-Retrieve (L2R)}, which repurposes the model's persistent internal state as a memory for historical context. A camera-conditioned retrieval gate selectively accesses relevant historical information from this state, determining \textit{what to retrieve}, while a retrieval trigger determines \textit{when to retrieve}. We further supervise the trigger with a 3D re-visibility signal, activating retrieval when previously observed content re-enters the current view while otherwise preserving the existing context. Together, these components enable the model to intrinsically acquire memory retrieval behavior and incorporate relevant historical observations into generation without a separate retrieval pathway. Across multiple base models and camera-revisit benchmarks, L2R improves long-term scene consistency while eliminating the need for an external memory bank or 3D conditions.
\end{abstract}


\section{Introduction}
\vspace{-3mm}

Video world models aim to generate video sequences that explore a 3D scene according to a user-specified camera trajectory, conditioned on one or a small set of reference frames \citep{yu2025contextmem,xiao2025worldmem,zhu2026sanawm}. Beyond perceptual realism and accurate camera control, a key requirement is scene-level consistency: upon revisiting previously observed regions, the generated content should remain consistent with earlier generations.

Existing strategies for enforcing scene consistency can be broadly grouped into two paradigms. \citet{yu2024viewcrafter,huang2025voyager,hu2026geometry,zheng2026versecrafter,yang2026neoverse} formulate this problem as conditional generation, leveraging reconstruction models~\citep{wang2025vggt,wang2026pi} to produce conditions intended to improve the fidelity and coherence of the generated video. However, the effectiveness of this approach is closely tied to reconstruction accuracy. Reconstruction errors can propagate to downstream generation, degrading long-term quality.

The second paradigm integrates memory retrieval into the video generation model to improve long-term scene consistency. In these methods, long-term memory serves as an auxiliary context module: when past frames are spatially relevant to the target view, they are retrieved to augment the default short-term sliding-window context. The principal differences between these methods lie in the retrieval strategy. Earlier approaches~\citep{li2025vmem,yu2025contextmem,xiao2025worldmem,yu2026mosaicmem,wang2026anchorweave} select historical context via camera overlap or explicit 3D signals and supply the retrieved views to the generator. More recent approaches make retrieval learnable, but still rely on separate memory access pathways. For example, MemLearner~\citep{yu2026memlearner} queries historical context using learnable query tokens, while CaR~\citep{peng2026compression} adds a dedicated retrieval-attention branch on compressed context. Although such designs improve long-range scene consistency, their reliance on external or auxiliary memory access pathways introduces a mismatch between retrieval and the model’s internal generative dynamics. 

In this paper, we advance the perspective that \textbf{\textit{video world models should intrinsically acquire the memory retrieval policy}}. Rather than treating memory retrieval as an external context construction procedure, we aim to make retrieval part of the generation process itself. This requires two capabilities to be jointly learned: maintaining historical information that remains accessible over long horizons, and selectively accessing that information according to the current observation. By integrating these two capabilities into the generative pathway, memory access can directly adapt to the model's evolving generation dynamics.

To realize this design, we formulate memory retrieval within an attention mechanism with a persistent contextual state and selective state access. In this formulation, historical observations are continuously accumulated in the model's internal state, while the state-access mechanism determines which historical information is retrieved for the current generation. Our construction builds on the state-update structure~\citep{katharopoulos2020linear,yang2024gla,yang2025gateddeltanet,pan2025scaling}, but repurposes it for memory retrieval. A retrieval trigger determines \textit{when to retrieve}, while a camera-conditioned gate determines \textit{what historical information to retrieve}, jointly coupling memory access with the current view. This shift repurposes the persistent attention state as an internal memory from which historical context can be selectively retrieved during long-horizon generation.

Building on this observation, we introduce \textbf{L2R}, a mechanism designed to approximate memory retrieval in video world models. L2R employs a camera-conditioned gate to selectively access historical information from the accumulated state according to the current viewpoint. In addition, L2R incorporates a retrieval trigger that determines whether a given token should initiate retrieval: when the current view revisits previously observed content, retrieval is activated; otherwise, the model relies on the existing context. We supervise the retrieval trigger using the 3D signal based on \textit{visibility re-entry}: tokens whose content exits the field of view or becomes occluded and subsequently reappears are annotated as retrieval-relevant. This supervision encourages the model to identify situations in which historical context is necessary. Collectively, L2R enables attention to acquire retrieval behavior intrinsically: the trigger learns \textit{when to retrieve}, and the camera-conditioned gate learns \textit{what to retrieve}, thereby coupling the current view with access to historical context without introducing a separate retrieval pathway or requiring explicit 3D correspondence at inference time.

We evaluate \method{} on scene exploration with re-appear and loop-closure trajectories under multiple state-update rules. Given a reference image and a camera trajectory, video world models using \method{} maintain scene consistency in the generated video without degrading visual quality or camera controllability. Both quantitative and qualitative results demonstrate that the method outperforms existing approaches in terms of consistency and exhibits robust generalization capabilities to unseen camera trajectories. Furthermore, ablation studies validate the effectiveness of the proposed design.


\section{Related Work}

\textbf{Memory in video generation.} In video generation, particularly in autoregressive (AR) settings, the incorporation of an explicit memory is critical for maintaining and exploiting long-term consistency. Existing systems represent and utilize historical information through a variety of mechanisms, including sliding-window strategies~\citep{chen2026context}, attention sinks~\citep{yang2025longlive}, positional extrapolation~\citep{zhao2025riflex,yesiltepe2026infinity}, state-space architecture~\citep{yu2025videossm}, token compression of past context~\citep{zhang2025packing}, and hybrid combinations of these techniques~\citep{yi2025deep}. More recently, ARL$^2$~\citep{li2026attend} introduced a memory mechanism for video generation, using the recurrent state as the memory bank. However, it lacks the ability to perform conditioned retrieval, thereby compromising the model's memory capabilities.

\textbf{Video world models} are designed to generate scene-level videos that are both temporally coherent and dynamically consistent, typically using video diffusion models conditioned on camera trajectories or action controls. In these scenarios involving camera pose or 3D data, memory can serve not only as an external component but can also be retrieved.
Many existing methods~\citep{yu2024viewcrafter,yu2025trajectorycrafter,hu2026geometry,yang2026neoverse,zheng2026versecrafter,wei2026geometry,xu2026ucm} incorporate substantial 3D information, such as depth maps, point clouds, or 3D Gaussian splatting (3DGS), as an additional conditioning memory for the diffusion process. A complementary research direction~\citep{li2025vmem,yu2025contextmem,xiao2025worldmem,wu2026video,xu2026wonder,zhu2026sanawm} investigates the memory retrieval and compression mechanisms that operate outside the diffusion backbone. These methods store and retrieve historical frames based on camera-view overlap~\citep{yu2025contextmem} or index past observations via geometric associations to observed three-dimensional surface segments~\citep{li2025vmem}. In addition, recent methods~\citep{peng2026compression,wang2026anchorweave,yu2026memlearner} introduce external learnable modules to retrieve memory for video world models, but keep retrieval separate from the generation model. Wonder~\citep{xu2026wonder} uses sparse attention, allowing inference to attend to a small subset of context tokens. However, its top-k selection offers no principled guarantee that the chosen tokens are the most geometrically relevant. 


\section{Method}
\label{sec:method}

In contrast to previous methods, we propose \textbf{\method{}}, a mechanism to accurately retrieve the historical context within the model, thus promoting improved consistency in the generated world videos. Sec.~\ref{sec:prelim} presents the necessary preliminaries. Sec.~\ref{sec:slots} specifies the implementation of our L2R and explains how it retrieves the historical context. Sec.~\ref{sec:train} summarizes the training pipeline.

\subsection{Preliminaries}
\label{sec:prelim}

\textbf{Memory retrieval in video world model.} Given a reference video $\{V_i\}_{i=1}^{M}$ and a novel camera trajectory $P_f$, a video world model $\varrho$ generates the video $V_f$ observed along the target trajectory. 

When the target viewpoint shares a high degree of consistency with a previous reference viewpoint (e.g. significant FoV overlap or similar geometric textures), existing approaches~\citep{yu2025contextmem} often introduce an external memory retrieval module to identify the trajectory-relevant information from the reference video and use the retrieved content as a context for generation.
\begin{equation}
\label{eq:external}
\mathcal{C}_f =
\begin{cases}
\mathrm{Retrieve}\!\left(\{V_i\}_{i=1}^{f-1}, P_f\right),
& \text{if target view revisits content}, \\[5pt]
\mathcal{C}_f^{\mathrm{base}},
& \text{otherwise},
\end{cases}
\qquad V_f = \varrho(\mathcal{C}_f, P_f),
\end{equation}
where $\mathcal{C}_f^{\mathrm{base}}$ denotes the default context used for generation.

Here, retrieval consists of two implicit decisions: whether historical information can be needed for the current view, and which historical information should be accessed when retrieval is activated. In existing methods, both decisions are made through an auxiliary memory access pathway outside the core generation process.

\subsection{Learning to retrieve}
\label{sec:slots}

\textbf{Internalize memory retrieval.} Inside the video diffusion transformer, video is compressed into a 1d token sequence $\{x_t\}_{t=1}^{N}$, where $t$ denotes the index of the token and $N$ is the sequence length. In this setting, Eq.~\ref{eq:external} will be transformed into the following form:

\begin{equation}
\label{eq:retrieval_state}
c_t = 
\begin{cases}
\mathrm{Retrieve}(\{x_i\}_{i=1}^{t-1}, p_t), & \text{if } r_t = 1, \\[5pt]
\{x_i\}_{i=t-w}^{t-1} & \text{if } r_t = 0,
\end{cases}
\qquad x_t = \varrho(c_t, p_t),
\end{equation}
where $p_t \in \mathbb{R}^{12}$ is the camera Plücker ray of $x_t$. $w$ is the size of the sliding window. $r_t$ indicates whether memory retrieval is activated for $x_t$. It emulates the video world model by initiating retrieval when the geometric overlap between the target and context views is sufficiently high. 

To internalize this retrieval, we store the historical context within the model state, allowing subsequent tokens to directly access the information from preceding tokens. A common practice is to encode such information in an internal recurrent state~\citep{yang2024gla,yang2025gateddeltanet,pan2025scaling,yu2025videossm}. Denoting the accumulated state before processing $x_t$ as $s_{t-1}$, which encodes information accumulated from preceding tokens, the historical context $c_t$ can therefore be read directly from this state.

To realize the retrieval defined in Eq.~\ref{eq:retrieval_state}, we read the historical context directly from the accumulated state. Specifically, $r_t$ determines whether retrieval is activated, while the pose-conditioned gate $\gamma_t$ determines which information is retrieved. The resulting context $c_t$ is then used to update the state:

\begin{equation}
\label{eq:slots}
c_t
=
\begin{cases}
{\gamma}_t \ s_{t-1}, & \text{If } r_t = 1, \\[6pt]
s_{t-1}, & \text{If } r_t = 0,
\end{cases}
\qquad s_t = \varphi(c_t, p_t),
\end{equation}
where $\varphi$ denotes the state transition rule.

\textbf{Designing $\color{red}\mathbf{{\gamma}_t}$: camera-guided retrieval.} As shown in Eq.~\ref{eq:slots}, memory retrieval relies on the camera pose $p_t$ as a condition. To this end, we introduce camera-guided retrieval for L2R:

\begin{equation}
\label{eq:pose}
\log \gamma_t \leftarrow \log \gamma_t \odot \big(1 + \tanh(W_g \phi_t)\big),
\end{equation}
where feature $\phi_t \in \mathbb{R}^{D_p}$ is the projection of $p_t$.

To promote numerical stability in the gated retrieval procedure, following previous work~\citep{yang2024gla,yang2025gateddeltanet,pan2025scaling}, we perform multiplicative gating computations in log space.

\textbf{Designing $\color{red}\mathbf{r_t}$: retrieval trigger and re-visibility supervision.} As shown in Eq.~\ref{eq:slots}, the behavior of L2R-GLA is consistent with the memory module of a video world model: the memory retrieval operator is triggered only when the viewpoint associated with the current token exhibits a high degree of consistency with viewpoints already represented in the context. The variable $r_t$ is introduced to parameterize and learn this retrieval activation.

\begin{figure}[t]
\centering
\includegraphics[width=\linewidth]{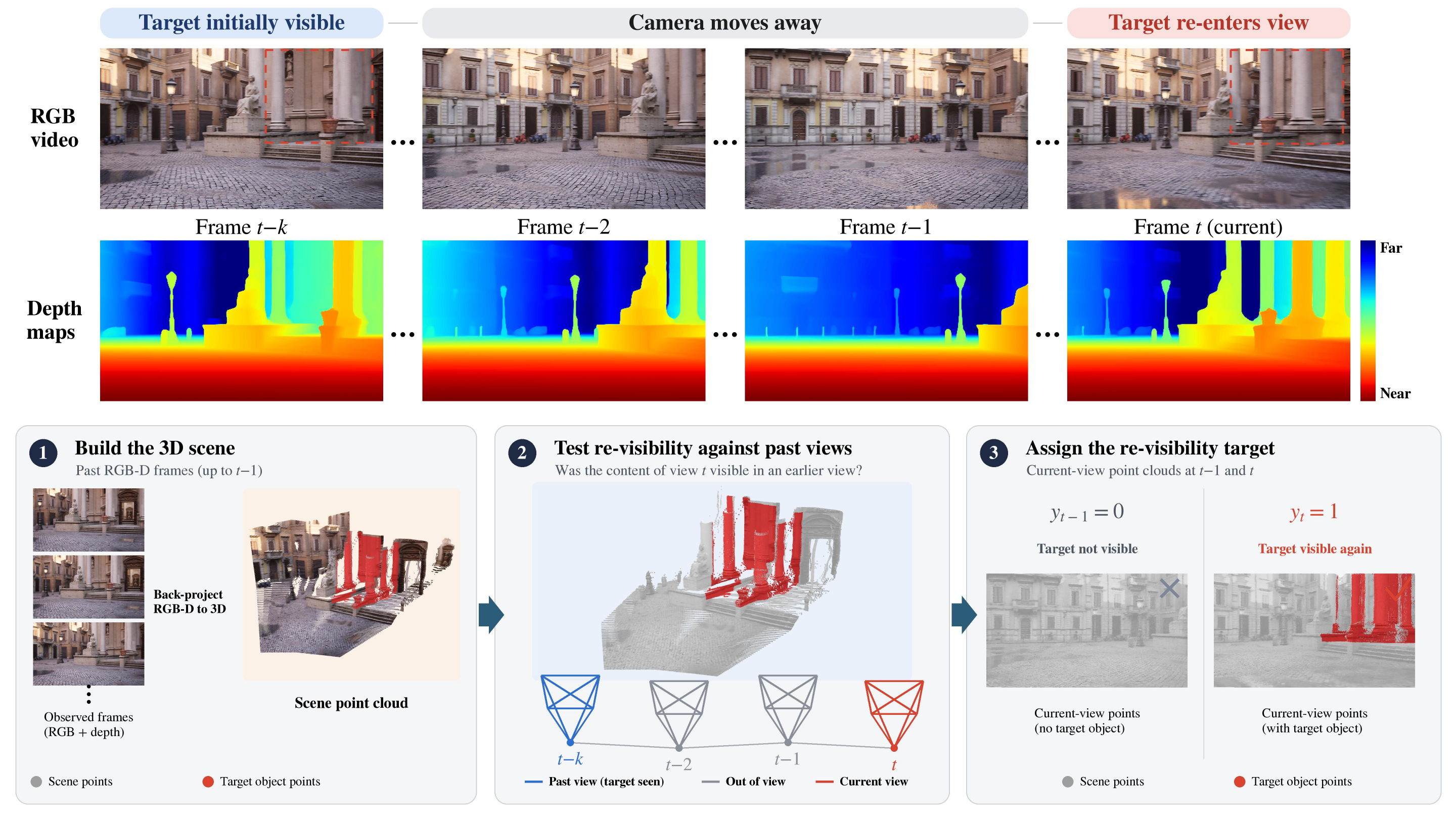}
\caption{\textbf{Three stages to compute re-visibility supervision.} (1) estimate per-frame depth; (2) unproject each token to 3D points; and (3) re-project previous frames' 3D points into the current frame. If a previously observed 3D point leaves the frustum (or is occluded) and later reappears, assign $y_t = 1$ to the corresponding token.}
\label{fig:qual}
\vspace{-3mm}
\end{figure}

We learn $r_t$ under re-visibility supervision. As shown in Figure~\ref{fig:qual}, for each frame in a training clip, the supervisory target for $r_t$ is computed offline via a three-stage procedure: (1) estimating per-frame depth using a feed-forward geometric model~\citep{wang2025vggt,wang2026pi}; (2) unprojecting each token in the frame to a 3D point using the estimated depth; and (3) re-projecting the 3D points of earlier frames into the current frame using the known camera trajectory. If a 3D surface point observed in the previous frame exits the frustum (or is occluded) and reappears in the current frame, the label $y_t = 1$ is assigned to the corresponding token. We optimize $r_t$ using the following objective: 

\begin{equation}
\label{eq:gateloss}
\mathcal{L}_{\mathrm{r}} = \frac{1}{|\mathcal{V}|} \sum_{t \in \mathcal{V}} \mathrm{BCE}_{w_+}\!\left( \hat r_t,\; y_t \right),
\end{equation}
where $\hat r_t$ is the sigmoid probability before binarization, $\mathcal{V}$ contains valid target-frame tokens, and $w_+=3$ is the positive-class weight. For numerical stability, we compute the weighted BCE directly from gate logits and mask invalid tokens and context frames. Since binarization is non-differentiable, we stop gradients through $r_t$ and force the model to retrieve memory on this discrete trigger. 

\textbf{Extension.} Eq.~\ref{eq:slots} defines a mechanism-independent retrieval recipe that can be instantiated with different gated state transition mechanisms~\citep{yang2024gla,yang2025gateddeltanet}. In this paper, we validate this by instantiating the same recipe with both Gated Linear Attention (GLA)~\citep{yang2024gla} and Gated DeltaNet (GDN)~\citep{yang2025gateddeltanet}. For GDN, the instantiation requires an additional modification to its delta correction, and we retain its subtractive memory gating. Details of this adaptation are provided in Appendix~\ref{sec:extension}.

Instantiating L2R with these two state transition rules yields \textbf{L2R-GLA} and \textbf{L2R-GDN}, respectively. We use these names throughout the remainder of this paper.

\subsection{Training}
\label{sec:train}

During model training, we support both training from scratch and supervised fine-tuning (SFT), depending on the experimental configuration required for evaluation. For most evaluation protocols, training is initialized from parameters loaded from a pre-trained video generation model.

\textbf{Integration into a pretrained model.} In every group of $n$ blocks in the pre-trained camera-controllable video diffusion model, we replace the first $n-1$ self-attention layers with our memory layer (L2R-GLA). The remaining blocks retain softmax attention, so the attention captures the local spatial structure while the memory state provides temporal persistence. For each replacement, we initialize the memory layer's query, key, value, and output projections from the attention projections, and zero-initialize all camera-dependent paths so the layer initially matches the pre-trained token representation. 

\textbf{Objective.} We optimize the rectified-flow objective together with re-visibility supervision:
\begin{equation}
\label{eq:objective}
\mathcal{L}
= \mathcal{L}_{\mathrm{denoise}}
+ \lambda_{\mathrm{r}}\,\mathcal{L}_{\mathrm{r}},
\end{equation}
where $\mathcal{L}_{\mathrm{denoise}}$ is the timestep-weighted mean-squared error between the predicted flow velocity and $\epsilon-z_0$, where $z_0$ denotes the clean video latents and $\epsilon$ is Gaussian noise. This loss is computed only on target-frame latents; reference and context frames are excluded. The visibility loss $\mathcal{L}_{\mathrm{r}}$ in Eq.~\ref{eq:gateloss} is normalized over valid target tokens within each memory layer and then averaged across memory layers. We use $\lambda_{\mathrm{r}}=0.05$ by default. 

\section{Experiments}
\label{sec:experiments}

\subsection{Setup}

We evaluate the performance of L2R-GLA and L2R-GDN along two dimensions: \textbf{\textit{memory capacity}} and \textbf{\textit{video quality}}. The evaluation of the former characterizes the module’s ability to maintain long-term consistency, whereas the evaluation of the latter verifies that this memory mechanism does not degrade visual fidelity or compromise camera controllability.

\textbf{Memory capacity.} Following Echo-Memory~\citep{echomem2026}, we conduct a comprehensive evaluation of memory capabilities. We adopt Wan2.1-T2V-1.3B~\citep{wan2025wan} as the base model and follow the camera-conditioned method introduced in Echo-Memory. To ensure experimental comparability, we use the static scene-exploration dataset~\citep{yu2025contextmem} and \textit{all methods use exactly the same training and inference settings}. The implementation details are shown in Appendix~\ref{sec:details}. Performance is evaluated using a three-branch protocol. (i) \emph{Replay}: the model generates videos conditioned on given camera trajectories, and memory capacity is quantified using frame-aligned Peak Signal-to-Noise Ratio (PSNR), Structural Similarity Index Measure (SSIM), and Learned Perceptual Image Patch Similarity (LPIPS). (ii) \emph{In-domain loop closure}: outgoing and returning frames at matched camera poses are compared using the same image-quality metrics to measure revisit consistency. (iii) \emph{Open domain return (O-V)}: scene-level consistency and object-identity preservation are assessed on first-frame probes released using the benchmark’s vision language judge. This assesses the model's generalization capability to unseen trajectories and scenes.

\textbf{Video quality.} To isolate whether our memory modules improve long-horizon coherence \emph{without} trading off perceptual fidelity, we curate a training corpus of 5-second clips \textit{without camera loops}, sampled from RealEstate10K, DL3DV, and MiraData9K. For evaluation, we build a ``forth-and-back'' set on RealEstate10K and DL3DV, in which the second half explicitly retraces the camera path back to the start, yielding a controlled test of \textit{out-of-domain} viewpoint regression under increasing drift. Visual quality is reported with VBench~\citep{huang2024vbench}. Camera controllability is measured by pose agreement between the ground-truth trajectory and the trajectory re-estimated from generated frames using Pi3X~\citep{wang2026pi}. We report rotation and translation errors $(R_{err}, T_{err})$ following VMem~\citep{li2025vmem}. We instantiate the pipeline on Wan2.2-TI2V-5B and SANA-WM-2.6B~\citep{zhu2026sanawm}. Remaining implementation details are deferred to Appendix~\ref{sec:details}.

\subsection{Results on memory capacity}
\label{sec:res_main}

Table~\ref{tab:main_echo} compares memory capacity across replay, loop closure, and open-domain return. \methodshort{}-GLA performs best, ranking first on all replay metrics and on SSIM and O-V for loop closure and open-domain return. On replay, it raises PSNR from 13.71 (best baseline, VMem~\citep{li2025vmem}) to 14.74 (+1.03 dB), with higher SSIM and lower LPIPS. Versus Context as Memory~\citep{yu2025contextmem} at $K{=}20$, it improves PSNR by 2.20 dB (12.54$\rightarrow$14.74), SSIM by 0.064 (0.449$\rightarrow$0.513), and reduces LPIPS by 0.066 (0.496$\rightarrow$0.430). The GLA baseline degrades in SSIM and LPIPS, likely because unconstrained retrieval causes memorization drift. Overall, integrating retrieval into the video generation model substantially improves consistency when previously observed content is revisited. L2R-GDN also shows strong performance, achieving 13.86 dB PSNR on replay and 69.37 O-V, which suggests that the proposed formulation remains effective when combined with GDN.

\newcommand{\ours}{\rowcolor{blue!6}}
\newcommand{\best}[1]{\textbf{#1}}
\newcommand{\second}[1]{\underline{#1}}
\newcommand{\pend}{\textcolor{black!35}{--}}

\begin{table}[t]
\caption{\textbf{Quantitative comparison on memory capacity.} Best in \best{bold}, second best \second{underlined}. $^*$ denotes methods are retrained because their released weights were not trained on dataset~\citep{yu2025contextmem}. The results for the other methods are taken from Echo-Memory~\citep{echomem2026}.}
\label{tab:main_echo}
\centering
\small
\renewcommand{\arraystretch}{1.1}
\setlength{\tabcolsep}{3pt}
\begin{tabularx}{\linewidth}{@{}l *{7}{Y}@{}}
\toprule
\multirow{2}{*}{\textbf{Method}} & \multicolumn{3}{c}{Replay} & \multicolumn{3}{c}{Loop closure} & \multirow{2}{*}{O-V $\uparrow$} \\
\cmidrule(lr){2-4} \cmidrule(lr){5-7}
 & PSNR $\uparrow$ & SSIM $\uparrow$ & LPIPS $\downarrow$ & PSNR $\uparrow$ & SSIM $\uparrow$ & LPIPS $\downarrow$ & \\
\midrule
I2V baseline (Wan2.1)                 & 10.03 & 0.398 & 0.534 & 10.32 & 0.291 & 0.643 & 12.25 \\
Spatial Memory               & 13.60 & 0.411 & 0.554 & 10.04 & 0.260 & 0.617 & 6.00 \\
VideoSSM  & 12.69 & 0.344 & 0.581 & 12.23 & 0.298 & 0.584 & 34.75 \\
ARL$^{2 \ *}$ & 12.73 & 0.326 & 0.594 & 12.13 & 0.309 & 0.633 & 43.06 \\
Context as Memory, $K{=}5$    & 11.92 & 0.408 & 0.501 & 10.72 & 0.307 & 0.596 & 50.75 \\
Context as Memory, $K{=}20$   & 12.54 & 0.449 & \second{0.496} & 11.07 & 0.359 & \best{0.543} & 58.63 \\
WorldMem$^{*}$   & 12.34 & 0.471 & 0.536 & 11.59 & 0.348 & 0.594 & 59.74 \\
VMem$^{*}$ & 13.71 & \second{0.502} & 0.507 & \best{13.17} & 0.352 & \second{0.581} & 64.73 \\
\midrule
\ours \textbf{\methodshort{}-GLA (ours)}  & \best{14.74} & \best{0.513} & \best{0.430} & \second{13.14} & \best{0.375} & 0.603 & \best{72.57} \\
\ours \textbf{L2R-GDN (ours)} & \second{13.86} & 0.409 & 0.558 & 12.81 & \second{0.374} & 0.614 & \second{69.37} \\
\bottomrule
\end{tabularx}
\vspace{-1mm}
\end{table}

\begin{figure}[t]
\centering
\vspace{-3mm}
\includegraphics[width=\linewidth]{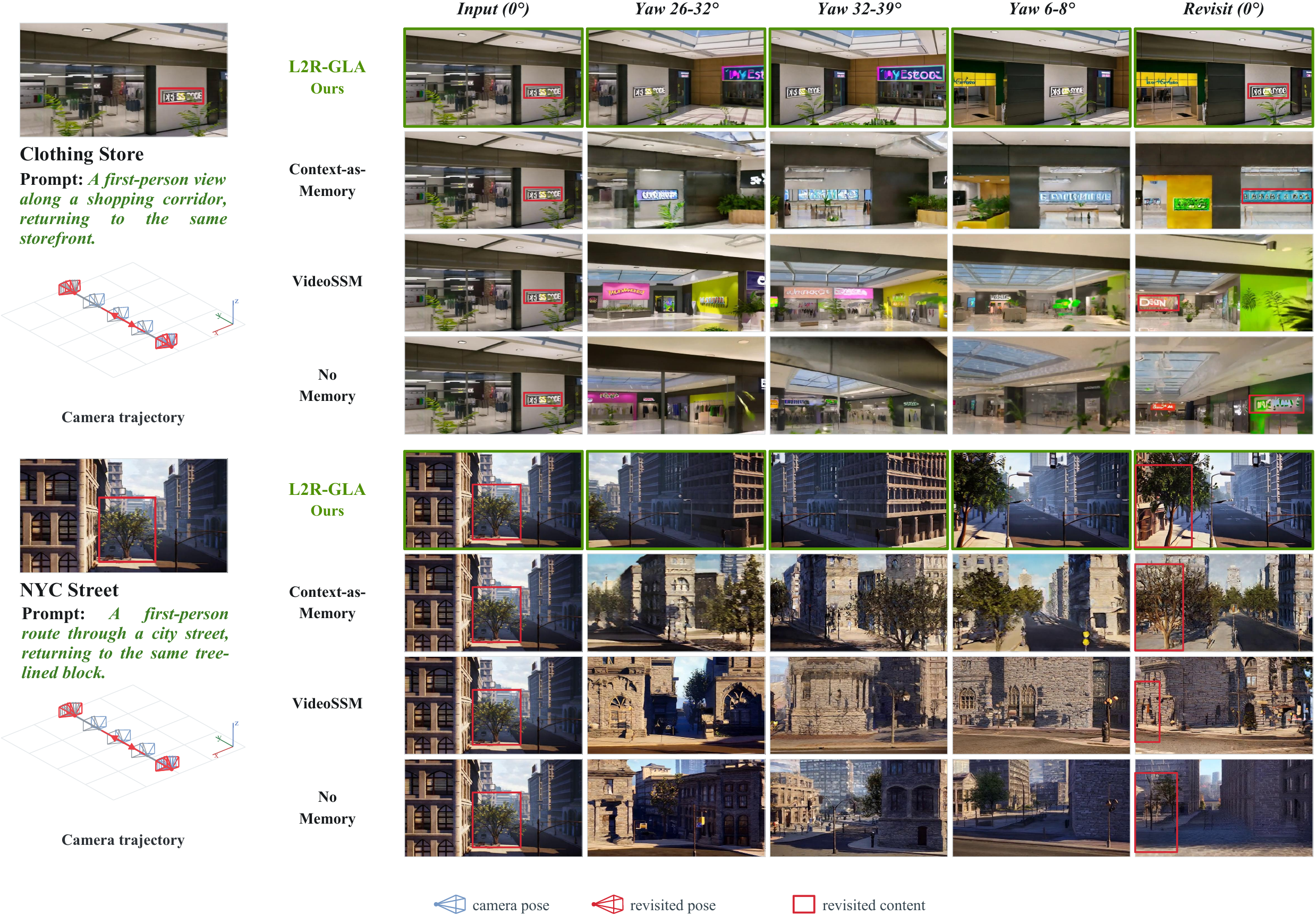}
\caption{\textbf{Qualitative comparison on memory capacity.} Red boxes highlight the revisited regions. \methodshort{}-GLA has the best scene consistency.}
\label{fig:qual_loop}
\vspace{-5mm}
\end{figure}

A similar trend is observed under loop closure. \methodshort{}-GLA achieves the best SSIM of 0.375 and the second-best PSNR of 13.14, closely matching the strongest baseline, VMem, at 13.17. It also substantially improves over Context as Memory at $K{=}20$, increasing PSNR from 11.07 to 13.14 and SSIM from 0.359 to 0.375. Although Context as Memory attains a lower LPIPS, \methodshort{}-GLA provides stronger pixel-level fidelity together with higher structural similarity. This indicates that the proposed internal retrieval mechanism is particularly effective at preserving the appearance and structure of previously observed content during viewpoint revisits.

For open-domain return, \methodshort{}-GLA achieves an O-V score of 72.57, outperforming VMem~\citep{li2025vmem} by 7.84 points and Context as Memory at $K{=}20$ by 13.94 points. Since this protocol evaluates held-out first-frame probes with distinctive objects, the improvement suggests that the learned internal retrieval policy can generalize beyond the training trajectories and recover previously observed scene content under unseen return scenarios. This further supports the ability of L2R to preserve scene identity without relying on an external memory bank.

The qualitative comparisons in Figure~\ref{fig:qual_loop} further illustrate the memory behavior of L2R-GLA. After the camera moves away from a region and later returns to a matched viewpoint, L2R-GLA better preserves the appearance of the revisited content while maintaining the surrounding scene layout. The red boxes highlight these regions, showing that internal retrieval helps recover previously observed details with improved scene consistency.

\subsection{Results on video quality}

Table~\ref{tab:re10k_dl3dv} demonstrates that L2R improves revisit consistency while maintaining strong video quality and camera controllability across different underlying video generation models. With Wan2.2 as the base model, both L2R variants achieve the highest or second-highest scores across all six visual-quality dimensions, with L2R-GLA ranking first in all of them. Meanwhile, L2R substantially improves camera controllability over prior memory-based methods. Compared with CaR, L2R-GLA reduces the rotation and translation errors from 12.98 to 6.29 and from 0.354 to 0.300, respectively, and L2R-GDN achieves 6.28 and 0.282. For revisit consistency, L2R-GDN attains the best PSNR of 17.03, improving over CaR by 0.19\,dB, while L2R-GLA remains competitive at 16.79. These results indicate that the proposed internal retrieval mechanism improves scene consistency without requiring a trade-off with video quality or camera control.

Using SANA-WM as the base model, the same trend is observed. L2R-GDN achieves substantial gains in revisit consistency, improving PSNR from 15.52 to 17.29 and SSIM from 0.5076 to 0.5597, while also reducing the translation error from 0.383 to 0.246. Although its visual-quality scores are slightly lower than the SANA-WM baseline, the gap remains small across the six dimensions. L2R-GLA similarly improves revisit consistency to 16.97 PSNR and 0.5556 SSIM while maintaining competitive visual quality and camera control. Overall, these results show that L2R transfers across different underlying state-transition mechanisms and consistently strengthens long-term scene consistency without substantially degrading generation quality or camera controllability.

\begin{table}[t]
\caption{\textbf{Quantitative results of video quality, camera controllability, and re-visit consistency} on RealEstate10K and DL3DV samples. Best in \best{bold}, second best \second{underlined}. $^*$ marks methods we retrained on our dataset because checkpoints weren’t released; others use official checkpoints.}
\label{tab:re10k_dl3dv}
\centering
\renewcommand{\arraystretch}{1.1}
\setlength{\tabcolsep}{3pt}
\resizebox{\linewidth}{!}{%
\begin{tabular}{lcccccccccc}
\toprule
\multirow{3}{*}{\textbf{Method}} &
\multicolumn{6}{c}{\textbf{Visual quality ($\uparrow$)}} &
\multicolumn{2}{c}{\textbf{Camera control ($\downarrow$)}} &
\multicolumn{2}{c}{\textbf{Consistency  ($\uparrow$)}} \\
\cmidrule(lr){2-7}
\cmidrule(lr){8-9}
\cmidrule(lr){10-11}
&
\textbf{Subject} &
\textbf{Bg} &
\textbf{Motion} &
\textbf{Temporal} &
\textbf{Aesthetic} &
\textbf{Imaging} &
\multirow{2}{*}{$R_{err}$} &
\multirow{2}{*}{$T_{err}$} &
\multirow{2}{*}{PSNR} &
\multirow{2}{*}{SSIM} \\
&
\textbf{Consist} &
\textbf{Consist} &
\textbf{Smooth} &
\textbf{Flicker} &
\textbf{Quality} &
\textbf{Quality} &
& \\
\midrule

ViewCrafter
& 78.51 & 88.30 & 95.12 & 92.43 & 46.28 & 63.42 & 8.86 & 0.482 & 13.81 & 0.5217 \\

SEVA
& 90.53 & 93.79 & 96.68 & 93.41 & 46.01 & 64.08 & \best{5.26} & \second{0.249} & 16.62 & 0.5427 \\

VMem
& 87.71 & 92.50 & 94.16 & 90.41 & 42.05 & 61.50 & 10.31 & 0.518 & 16.85 & 0.5505  \\

SANA-WM & 91.73 & 93.67 & 96.81 & 93.99 & 46.58 & 66.47 & \second{5.40} & 0.383 & 15.52 & 0.5076 \\

AnchorWeave-Wan2.2$^{*}$ 
& 88.70 & 92.35 & 95.94 & 92.71 & 47.50 & 65.24 & 6.81 & 0.299 & 16.55 & 0.5472 \\

CaR-Wan2.2 & 91.47 & 93.28 & 97.56 & 94.37 & 49.20 & \second{66.34} & 12.98 & 0.354 & 16.84 & 0.5456 \\

\midrule

\ours \textbf{L2R-GLA (Wan2.2)}
& \best{92.67} & \best{94.41} & \best{97.68} & \best{95.53} & \best{50.14} & \best{67.18} & 6.29 & 0.300 & 16.79 & 0.5480 \\

\ours \textbf{L2R-GDN (Wan2.2)}
& \second{92.46} & \second{94.38} & \second{97.62} & \second{95.24} & \second{49.60} & {65.97} & 6.28 & 0.282 & \second{17.03} & 0.5546 \\

\ours \textbf{L2R-GLA (SANA-WM)}
& 91.36 & 94.17 & 97.02 & 93.84 & 46.81 & 66.07 & 5.61 & 0.259 & 16.97 & \second{0.5556} \\

\ours \textbf{L2R-GDN (SANA-WM)}
& 90.85 & 93.86 & 96.68 & 93.43 & 46.16 & 65.13 & \second{5.40} & \best{0.246} & \best{17.29} & \best{0.5597} \\

\bottomrule
\end{tabular}%
}
\vspace{-3mm}
\end{table}

\subsection{Ablation}
\label{sec:res_ablation}

To fully understand how each design element in L2R-GLA affects its memory capability, we conducted ablation studies on Echo-Memory.

\begin{table}[htbp]
\vspace{-2mm}
\caption{\textbf{Ablation on the memory capacity.} GLA (baseline) is our starting point. The last three rows progressively add camera-guided retrieval $\gamma_t$, retrieval trigger $r_t$ and re-visibility supervision (Eq.~\ref{eq:gateloss}). Best in \best{bold}, second best \second{underlined}.}
\label{tab:abl_echo}
\vspace{-1mm}
\centering
\small
\renewcommand{\arraystretch}{1.1}
\setlength{\tabcolsep}{3pt}
\begin{tabularx}{\linewidth}{@{}l *{7}{Y}@{}}
\toprule
\multirow{2}{*}{\textbf{Variant}} & \multicolumn{3}{c}{Replay} & \multicolumn{3}{c}{Loop closure} & \multirow{2}{*}{O-V $\uparrow$} \\
\cmidrule(lr){2-4} \cmidrule(lr){5-7}
 & PSNR $\uparrow$ & SSIM $\uparrow$ & LPIPS $\downarrow$ & PSNR $\uparrow$ & SSIM $\uparrow$ & LPIPS $\downarrow$ & \\
\midrule
No Memory                         & 10.03 & 0.398 & 0.534 & 10.32 & 0.291 & 0.643 & 12.25 \\
VideoSSM                          & 12.69 & 0.344 & 0.581 & 12.23 & 0.298 & \best{0.584} & 34.75 \\
Context as Memory, $K{=}5$          & 11.92 & 0.408 & 0.501 & 10.72 & 0.307 & 0.596 & 50.75 \\
\midrule
GLA                    & 10.55 & 0.199 & 0.763 & 12.74 & 0.273 & \second{0.584} & 66.13 \\
+ Camera-guided retrieval ($\gamma_t$)          & 13.88 & \second{0.421} & 0.560 & 12.76 & \best{0.399} & 0.687 & 68.74 \\
+ Retrieval trigger ($r_t$) & \second{14.02} & 0.402 & \second{0.483} & \second{13.11} & \second{0.383} & 0.601 & \second{71.42} \\
+ Re-visibility supervision (Eq.~\ref{eq:gateloss}) & \best{14.74} & \best{0.513} & \best{0.430} & \best{13.14} & 0.375 & 0.603 & \best{72.57} \\
\bottomrule
\end{tabularx}
\vspace{-2mm}
\end{table}

\begin{figure}[t]
  \centering
  \begin{minipage}[t][0.33\textwidth][t]{0.36\textwidth}
    \vspace{0pt}
    \centering
    \begingroup
    \small
    \renewcommand{\arraystretch}{1.1}
    \setlength{\tabcolsep}{3pt}
    \resizebox{0.95\linewidth}{!}{%
      \begin{tabular}{@{}lccc@{}}
        \toprule
        \multirow{2}{*}{\textbf{Method}}
          & \multirow{2}{*}{\shortstack{\textbf{Layer}\\\textbf{ratio}}}
          & \textbf{Replay} & \textbf{Loop closure} \\
          & & PSNR $\uparrow$ & PSNR $\uparrow$ \\
        \midrule
         I2V baseline & - & 10.03 & 10.32 \\
        \midrule
        \multirow{3}{*}{L2R-GDN}
          & $1{:}1$ & 13.62 & 12.98 \\
          & $3{:}1$ & 13.88 & 12.76 \\
          & $7{:}1$ & 8.95 & 8.75 \\
        \midrule
        \multirow{3}{*}{L2R-GLA}
          & $1{:}1$ & 14.26 & 13.55\\
          & $3{:}1$ & 14.74  & 13.14 \\
          & $7{:}1$ & 10.52 & 10.37 \\
        \bottomrule
      \end{tabular}%
    }
    \endgroup
    \vfill
    {\small\textbf{(a) Layer-ratio ablation}\par}
  \end{minipage}\hfill
  \begin{minipage}[t][0.33\textwidth][t]{0.62\textwidth}
    \vspace{0pt}
    \centering
  \includegraphics[width=\linewidth]{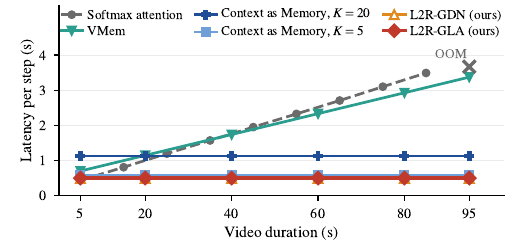}
    \vfill
    {\small\textbf{(b) Latency per denoising step}\par}
  \end{minipage}
\vspace{-1mm}
\caption{\textbf{Layer-ratio ablation and inference efficiency.} Left: L2R-GDN and L2R-GLA with layer ratios of $1{:}1$, $3{:}1$, and $7{:}1$ on Echo-Memory benchmark, evaluated by replay and loop-closure PSNR.
 Right: latency comparison on one H100, $5$\,s chunks.}
\label{fig:layers}
\vspace{-5mm}
\end{figure}

\textbf{Effects of the retrieval gate $\gamma_t$ and the retrieval trigger $r_t$.} Table~\ref{tab:abl_echo} separates the two mechanisms and highlights their distinct contributions. Starting from the GLA baseline, introducing camera-guided retrieval with $\gamma_t$ yields the largest improvement in replay quality, with PSNR increasing from 10.55 to 13.88, SSIM from 0.199 to 0.421, and LPIPS decreasing from 0.763 to 0.560. This demonstrates the importance of using the camera pose to selectively access relevant historical information. Adding the retrieval trigger $r_t$ further improves the results, increasing Replay PSNR from 13.88 to 14.02 and reducing LPIPS from 0.560 to 0.483, while also improving Loop-closure PSNR from 12.76 to 13.11 and LPIPS from 0.687 to 0.601. The overall O-V score likewise increases from 68.74 to 71.42. These results indicate that camera-conditioned retrieval provides the main capacity gain, while the retrieval trigger further improves retrieval selectivity by activating memory access when historical information is needed.

\textbf{Effects of the re-visibility supervision (Eq.~\ref{eq:gateloss}).} Re-visibility supervision further strengthens scene consistency by sharpening the trigger on revisits (Table~\ref{tab:abl_echo}). On replay, it improves PSNR by 0.72 dB, increases SSIM by 0.111, and reduces LPIPS from 0.483 to 0.430. Importantly, the auxiliary target is not merely an across-the-board image-metric regularizer. Its primary effect is to make triggering more selective and useful for revisited content, which directly translates into stronger memory.

\textbf{Layer ratios.} We further vary the ratio of L2R-GLA / L2R-GDN layers to standard softmax attention (Table~\ref{fig:layers} (a)). Both attention variants outperform the I2V baseline in replay and loop-closure PSNR, but the ratio matters: $3{:}1$ maximizes replay PSNR, $1{:}1$ maximizes loop-closure PSNR, and $7{:}1$ degrades both. We therefore adopt $3{:}1$ L2R-GLA in the main comparison to prioritize replay strength while keeping loop closure competitive. We attribute this trend to the fixed-capacity bottleneck of the linear-attention state: as trajectories grow longer and more diverse, the recurrent summary must overwrite earlier details, so stacking too many L2R-GLA layers amplifies information loss and can hurt both replay and loop closure. Retaining a fraction of softmax-attention layers acts as a high-bandwidth fallback that re-injects global context and stabilizes long-horizon generation, preventing this form of memory collapse. We discuss this further in Appendix~\ref{sec:fail}.

\subsection{Efficiency} 

Figure~\ref{fig:layers}(b) shows the per-step denoising cost on one H100 as rollout length increases, with resolution, camera conditioning, and chunk size matched. Full-history softmax attention rises from $0.43$\,s at $5$\,s to $3.67$\,s at $90$\,s and runs out of $80$\,GiB at $95$\,s. Both L2R layers are duration-invariant: $0.50$\,s for L2R-GLA and $0.49$\,s for L2R-GDN. External retrieval also decouples cost from duration but adds overhead: Context as Memory takes $0.59$\,s at $K{=}5$ and $1.13$\,s at $K{=}20$. Geometric retrieval pays to maintain memory: VMem rebuilds its surfel index each step by running point-map and global alignment over all stored frames, increasing from $0.70$\,s to $3.38$\,s between $5$\,s and $95$\,s chunks, near the softmax-attention curve.

\section{Conclusion}

This study establishes a useful analogy between external context memory retrieval mechanisms employed in scene-consistent video world models and the state-update dynamics. Leveraging this relationship, we introduce \method{}, which designs camera-guided retrieval, retrieval triggering, and re-visibility supervision through a 3D geometry-based target. The resulting module can be integrated into pretrained camera-controllable video world models. Empirical results indicate that \method{} substantially improves long-range scene consistency in video world models, eliminating the need for external memory retrievers or explicit 3D condition injection.

\subsection*{Ethics Statement} This paper presents research aimed at advancing the field of video world models. Our work may have multifaceted impacts on society. Given the rapidly growing capabilities of video generation technology, we urge against its misuse, such as the creation of misleading videos or video generation for malicious purposes.

\subsection*{Reproducibility Statement} We state that L2R is highly reproducible. Implementation and datasets details on our main experiences are provided in Appendix~\ref{sec:details}. It is anticipated that these descriptions can sufficiently demonstrate the reproducibility of L2R. We plan to open-source the code and weight files after the paper passes peer review.

\bibliography{iclr2027_conference}
\bibliographystyle{iclr2027_conference}

\newpage
\appendix

\section{Implementation details}\label{sec:details}

\subsection{More efficient analysis}

\textbf{Parameters.} The L2R layer reuses the query, key, value, and output projections of the self-attention it replaces, so the new parameters are projection layers to obtain the retrieval operator $\gamma_t$ and the retrieval trigger $r_t$. When using Wan2.1 as backbone, one softmax self-attention holds $9.45$\,M parameters and L2R-GLA holds $11.20$\,M, of which $9.45$\,M are inherited. Replacing the first three blocks of four blocks therefore adds $5.25$\,M parameters.

\textbf{FLOPs.} For one $5$\,s chunk of Wan2.1 ($N{=}18{,}480$, $D{=}1536$), a softmax self-attention layer costs $2.45$\,TFLOPs per forward pass, whereas L2R-GLA costs $0.49$\,TFLOPs, about $5\times$ fewer. L2R-GDN has the same linear-in-$N$ complexity, with its additional delta correction introducing only a constant-factor overhead over L2R-GLA. Replacing the first three blocks of four blocks therefore reduces the total cost of the $30$-block DiT from $111$ to $95$\,TFLOPs per forward pass ($14\%$). Unlike softmax attention, whose cost grows quadratically with sequence length, both L2R-GLA and L2R-GDN scale linearly with the context length, making the computational advantage more pronounced for longer contexts.

\subsection{Re-visit benchmark and evaluation on RealEstate10K and DL3DV} 
\label{app:revisit_bench}

We construct \textbf{Revisit-Bench}, a controlled benchmark for evaluating memory and loop-closure consistency in video world models. The benchmark is constructed from real camera trajectories and video frames from RealEstate10K and DL3DV-10K. All benchmark generation is deterministic with a fixed random seed of 42.

\subsubsection{Overview}
\label{app:revisit_overview}

A video world model should preserve sufficiently long-range scene information to maintain consistency when the camera moves away from an observed region and subsequently revisits it. In particular, when the camera returns to its initial viewpoint, the generated frame should be consistent with the initial observation. Revisit-Bench quantifies this capability through a controlled trajectory construction protocol.

Each benchmark sequence contains 81 frames, consisting of one reference frame and 80 generated target frames. The camera trajectory is derived from real relative camera motion and transformed into a temporal palindrome, such that the camera exactly returns to its initial viewpoint at a predefined revisit time. Under the static-scene assumption, every target viewpoint has a corresponding ground-truth frame from the original video, allowing direct pixel-level evaluation.

We construct two trajectory splits using the same set of 300 scenes, with different revisit patterns:
\begin{itemize}
    \item \texttt{palindrome\_far}: the folding point is 40, and the camera revisits the initial viewpoint at $t=80$. The camera traverses the entire trajectory before returning, resulting in the largest memory span.
    \item \texttt{palindrome\_near}: the folding point is 24, and the camera revisits the initial viewpoint at $t=48$. After revisiting, the camera continues moving in the original forward direction for 32 additional steps, preventing degenerate solutions in which the model simply remains at the revisited viewpoint.
\end{itemize}

Each split contains 300 scenes, including 150 scenes from RealEstate10K and 150 from DL3DV-10K. Therefore, each evaluated method produces 600 sequences in total across the two splits.

\subsubsection{Data Sources}
\label{app:revisit_data}

The benchmark is constructed from RealCam-Vid~\citep{zheng2025realcam}. We use video clips from {RealEstate10K} and {DL3DV-10K}.

Each entry in \texttt{RealCam-Vid\_train.npz} contains the following fields:
\begin{itemize}
    \item \texttt{dataset\_source}: either \texttt{RealEstate10K} or \texttt{DL3DV-10K};
    \item \texttt{video\_path}: the video path relative to \texttt{DATA\_ROOT};
    \item \texttt{camera\_extrinsics}: a sequence of world-to-camera (w2c) extrinsic matrices of shape $(F,4,4)$;
    \item \texttt{camera\_intrinsics}: normalized camera intrinsics $(f_x,f_y,c_x,c_y)$, where the focal lengths and principal-point coordinates are normalized by the source image width and height;
    \item \texttt{align\_factor}: a per-window scale factor used to normalize the translation magnitude, consistent with the training data loader;
    \item \texttt{short\_caption} and \texttt{long\_caption}: textual descriptions of the scene.
\end{itemize}

The benchmark is sampled from these data while using different trajectory constructions. Thus, the benchmark is designed to evaluate revisit consistency.

\subsubsection{Filtering}
\label{app:revisit_selection}

We randomly permute the source records using a fixed seed of 42 and scan them sequentially. For each source dataset, we select 150 scenes subject to the following criteria:
\begin{enumerate}
    \item \textbf{Video-level deduplication.} At most one window is selected from each source video.
    \item \textbf{Valid camera poses.} All extrinsic matrices and relative poses must contain finite values, and the required matrix inversions must be numerically stable.
    \item \textbf{Motion filtering.} Motion statistics are computed from the relative camera poses after applying the same alignment scale used during training. The median per-frame translation step must lie in $[0.02,0.8]$, removing nearly static clips and extreme trajectories, and the maximum single-frame rotation must not exceed $10^\circ$.
\end{enumerate}

\subsubsection{Camera Pose Processing}
\label{app:revisit_pose}

We process camera poses using the same coordinate transformation as the training data loader. Given the sampled source extrinsics $\mathbf{W}_{t}\in\mathbb{R}^{4\times4}$, which follow the OpenCV world-to-camera convention, we first convert them to camera-to-world coordinates and apply the per-window alignment factor:
\begin{equation}
    \mathbf{C}_{t}
    =
    \mathbf{W}_{t}^{-1},
    \qquad
    \mathbf{C}_{t}[:3,3]
    \leftarrow
    \mathbf{C}_{t}[:3,3]\cdot s,
\end{equation}
where $s$ denotes \texttt{align\_factor}. We then convert the poses back to world-to-camera coordinates and express them relative to the first frame:
\begin{equation}
    \mathbf{W}^{\mathrm{rel}}_{t}
    =
    \mathbf{W}_{t}
    \left(\mathbf{W}_{0}\right)^{-1},
    \qquad
    \mathbf{C}^{\mathrm{rel}}_{t}
    =
    \left(\mathbf{W}^{\mathrm{rel}}_{t}\right)^{-1}.
\end{equation}
Consequently, $\mathbf{C}^{\mathrm{rel}}_{0}=\mathbf{I}$.

The resulting poses use the OpenCV camera-to-world convention, where the camera coordinate axes are $x$-right, $y$-down, and $z$-forward. The median translation step is approximately 0.48 for the training clips and 0.17 for the benchmark sequences.

For intrinsics, we recover source-pixel camera intrinsics from the normalized intrinsics provided by RealCam-Vid. Given normalized parameters $(f_x,f_y,c_x,c_y)$ and source resolution $(w,h)$, the intrinsic matrix is
\begin{equation}
\mathbf{K}=
\begin{bmatrix}
f_xw & 0 & c_xw\\
0 & f_yh & c_yh\\
0 & 0 & 1
\end{bmatrix}.
\label{eq:revisit_intrinsics}
\end{equation}

We use $(w,h)=(1280,740)$ for RealEstate10K and $(w,h)=(960,580)$ for DL3DV-10K. The same intrinsic matrix is used for all 81 frames within each sequence and is stored as an $(81,3,3)$ array.

\subsubsection{Palindrome Trajectory Construction}
\label{app:revisit_trajectory}

Let $t\in\{0,\ldots,80\}$ denote the frame index, let $f$ denote the folding point, and let $\mathbf{C}^{\mathrm{rel}}_t$ denote the real relative camera poses obtained above. We construct the camera trajectory as
\begin{equation}
\mathbf{P}(t)=
\begin{cases}
\mathbf{C}^{\mathrm{rel}}_{\min(t,\,2f-t)},
& t\leq 2f,\\[4pt]
\mathbf{C}^{\mathrm{rel}}_{t-2f},
& t>2f \quad \text{(near split only)}.
\end{cases}
\label{eq:revisit_palindrome}
\end{equation}

The first part of the trajectory therefore mirrors the original real camera motion directly rather than interpolating poses or manually reversing motion directions. Every pose in the palindrome is consequently an actual viewpoint present in the source trajectory.

At $t=2f$, the camera pose is exactly the identity:
\begin{equation}
    \mathbf{P}(2f)=\mathbf{I}.
\end{equation}
This property is explicitly verified during benchmark construction with an absolute tolerance of $10^{-9}$.

For \texttt{palindrome\_far}, $f=40$, yielding the trajectory
\[
0\rightarrow40\rightarrow80,
\]
where the camera returns to the initial viewpoint at $t=80$. There is no post-revisit segment.

For \texttt{palindrome\_near}, $f=24$, yielding
\[
0\rightarrow24\rightarrow48,
\]
followed by
\[
t=49,\ldots,80
\quad\mapsto\quad
\mathbf{C}^{\mathrm{rel}}_{1},\ldots,\mathbf{C}^{\mathrm{rel}}_{32}.
\]
Thus, after returning to the initial viewpoint at $t=48$, the camera continues along the original forward direction for 32 additional steps. This post-revisit segment prevents a trivial solution in which the generated trajectory remains stationary after loop closure.

\subsubsection{Ground-Truth Frames and Static-Scene Assumption}
\label{app:revisit_gt}

We assume a static scene, following the standard setting for loop-closure evaluation. Under this assumption, the ground-truth image corresponding to each generated viewpoint is the source frame associated with that viewpoint. We therefore define
\begin{equation}
\mathrm{gt\_frame\_indices}[t]=
\begin{cases}
\min(t,\,2f-t),
& t\leq2f,\\[4pt]
-(t-2f),
& t>2f \quad \text{(near split only)}.
\end{cases}
\label{eq:revisit_gt_index}
\end{equation}
The negative indices in the near split identify the post-revisit overshoot segment and indicate that the absolute value should be used to retrieve the corresponding source frame. In particular, frames $t=0$ and $t=2f$ correspond to exactly the same source viewpoint. Hence, the ground-truth frame at the revisit point is identical to the input reference image. 

\subsubsection{Text Prompts}
\label{app:revisit_prompt}

The text prompt for each sequence is taken from the \texttt{short\_caption} field of RealCam-Vid. If the short caption is unavailable, we fall back to \texttt{long\_caption}; if both captions are missing, we use the fixed prompt ``A slow camera motion through a real scene.'' The resulting prompts contain approximately 10--25 words, with an average length of 18.2 words.

\subsubsection{Statistics}
\label{app:revisit_statistics}

Each split contains 300 scenes, consisting of 150 RealEstate10K scenes and 150 DL3DV-10K scenes. Since both splits use the same set of source scenes, each evaluated method produces 600 generated sequences in total: 300 for the far split and 300 for the near split.

For the original trajectories before palindrome construction, the median per-frame translation step, measured in the aligned coordinate scale, has a mean of 0.408 and a median of 0.171, with values ranging from 0.040 to 0.783. The maximum per-frame translation step has a mean of 0.39, while the maximum single-frame rotation has a mean of $2.19^\circ$. The latter is bounded by the $10^\circ$ filtering threshold. For comparison, the median translation step of the training clips is approximately 0.48.

\subsubsection{Evaluation Protocol}
\label{app:revisit_metrics}

We evaluate generated videos using six groups of metrics following \cite{li2025vmem,huang2024vbench,wang2026dit,hu2026omni}.

\paragraph{Revisit consistency.}
The primary metric measures the agreement between the generated revisit frame $t_r=2f$ and the input reference image. We report PSNR and SSIM, with LPIPS optionally reported as an additional perceptual metric. For uint8 images, PSNR is computed as
\begin{equation}
\mathrm{PSNR}
=
10\log_{10}
\left(
\frac{255^2}{\mathrm{MSE}}
\right).
\end{equation}
SSIM is computed on grayscale images after averaging the three RGB channels, using $\mathrm{data\_range}=255$.

\paragraph{Camera accuracy.}
We additionally follow the VMem~\citep{li2025vmem} evaluation protocol for camera trajectory accuracy. Camera poses are estimated from the generated video using Pi3X and compared with the ground-truth camera poses after applying the same alignment procedure. Both trajectories are first expressed relative to their first frame:
\begin{equation}
\mathbf{P}'(t)
=
\mathbf{P}(0)^{-1}\mathbf{P}(t).
\end{equation}
This removes the arbitrary global origin and orientation associated with the estimated trajectory.

Because the estimated camera trajectory contains an unknown global similarity scale, translation is normalized by the distance of the farthest frame:
\begin{equation}
\mathbf{t}
\leftarrow
\frac{\mathbf{t}}
{\max_i\|\mathbf{t}_i\|_2}.
\end{equation}
No Umeyama alignment is applied. We then report the mean translation error
\begin{equation}
d_t
=
\left\|
\mathbf{t}_{\mathrm{gt}}-\mathbf{t}_{\mathrm{gen}}
\right\|_2
\end{equation}
over $t=1,\ldots,T-1$, together with the rotational geodesic error
\begin{equation}
d_R
=
\arccos
\left(
\frac{
\operatorname{tr}
\left(
\mathbf{R}_{\mathrm{gen}}
\mathbf{R}_{\mathrm{gt}}^{\mathsf T}
\right)-1
}{2}
\right),
\end{equation}
reported in degrees.

\paragraph{VBench.}
We further report six VBench~\cite{huang2024vbench} dimensions: Subject Consistency (DINO), Background Consistency (CLIP), Motion Smoothness (AMT-S), Temporal Flickering (frame-difference MAE), Aesthetic Quality (CLIP with a linear head), and Imaging Quality (MUSIQ). The first 81 frames of each generated video are used, and the prompt recorded in the corresponding manifest entry is provided as the text input.

\subsection{Training Settings}

\textbf{Memory capacity.} Following Echo-Memory~\citep{echomem2026}, we train L2R-GLA and L2R-GDN on 8 A100 GPUs. The hyper-parameters are as follows.

\begin{table}[ht]
\centering
\small
\begin{tabular}{ll}
\toprule
\textbf{Setting} & \textbf{Value} \\
\midrule
Backbone                                & Wan2.1 \\
Resolution                              & $352\times 640$ \\
Chunk length $T$                      & $81$ frames \\
Optimizer                               & AdamW \\
Learning rate                           & $5\!\times\!10^{-5}$ \\
Per-device batch / grad accumulation    & $1$ / $1$ \\
GPUs                                    & $8$ (A100-80G) \\
Total steps                             & $5\,$k \\
Timestep shift                          & $15$ \\
Spike rejection threshold               & $15.0$ \\
Target-frame-only supervision           & enabled \\
Flow noise shift                         & $1.0$ \\
Relative-RT action encoding              & enabled \\
\bottomrule
\end{tabular}
\end{table}

\textbf{Video quality.} Although we use two base models, we consult the same training hyperparameters for them to train our own model.

\begin{table}[ht]
\centering
\small
\begin{tabular}{ll}
\toprule
\textbf{Setting} & \textbf{Value} \\
\midrule
Backbone                                & Wan2.2 / SANA-WM \\
Resolution                              & $704\times 1280$ \\
Chunk length $T$                      & $81$ frames \\
Optimizer                               & AdamW \\
Learning rate                           & $8\!\times\!10^{-5}$ \\
Per-device batch / grad accumulation    & $2$ / $1$ \\
GPUs                                    & $64$ (H20-96G) \\
Total steps                             & $5\,$k \\
Timestep shift                          & $9.98$ \\
Spike rejection threshold               & $1.0$ \\
Target-frame-only supervision           & enabled \\
Flow noise shift                         & $1.0$ \\
Relative-RT action encoding              & enabled \\
\bottomrule
\end{tabular}
\end{table}

\section{Extension to other state update rules}\label{sec:extension}

L2R defines the retrieval condition $r_t$ and the state-access gate $\gamma_t$, while the underlying state-space mechanism determines how the retrieved state is updated. In implementation, following~\cite{yang2024gla,yang2025gateddeltanet,pan2025scaling,qiu2026hydra,zhang2026hydra-x}, the vector-valued retrieval gate $\gamma_t$ is converted to a diagonal matrix.

\begin{equation}
S_t=
\begin{cases}
\gamma_t \ S_{t-1}+v_tk_t^\top, & \text{If } r_t=1,\\
S_{t-1}+v_tk_t^\top, & \text{If } r_t=0.
\end{cases}
\end{equation}

For GDN, the same retrieval formulation is combined with its delta correction:
\begin{equation}
S_t=
\begin{cases}
\gamma_t \ \!\left(S_{t-1}-\beta_t S_{t-1}k_tk_t^\top\right)
+\beta_t v_tk_t^\top, & \text{If } r_t=1,\\
S_{t-1}+\beta_t v_tk_t^\top, & \text{If } r_t=0.
\end{cases}
\end{equation}
Here, $\gamma_t$ controls which historical information is accessed, while $\beta_t$ controls the subtractive delta update and corresponding write strength in GDN.


\section{Failure mode}\label{sec:fail}
\begin{figure}[t]
\centering
\includegraphics[width=\linewidth]{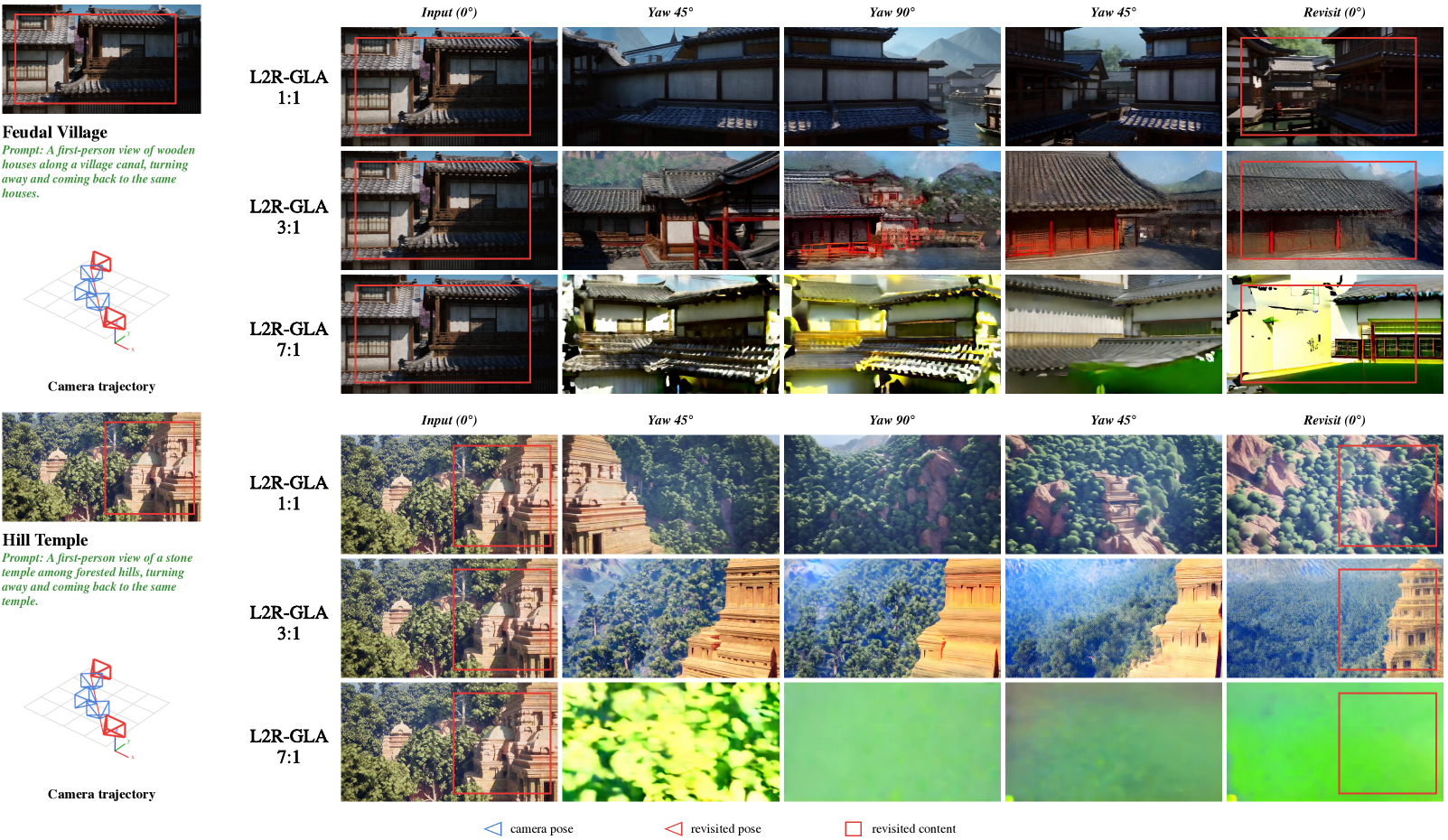}
\caption{Failure mode at a high memory-layer ratio. Loop closure on two scenes with memory-layer ratios of $3{:}1$, $1{:}1$ and $7{:}1$. With too few memory layers the rollout loses the observed structure on the return.}
\label{fig:failure}
\vspace{-3mm}
\end{figure}

Figure~\ref{fig:failure} shows loop-closure rollouts on the same scenes for the two retrieval baselines and for \methodshort{}-GLA with a $7{:}1$ memory-layer ratio. With only four memory layers, the recurrent state has to summarize the whole trajectory at few depths of the network, and the softmax layers that remain no longer provide a high-bandwidth fallback for the content the state has overwritten: the return frames lose the observed structure and the loop is not closed, in line with the drop of the $7{:}1$ rows in Figure~\ref{fig:layers}(a), whereas the $1{:}1$ and $3{:}1$ ratios close the loop (Section~\ref{sec:res_ablation}).

\section{Future work}

Future research could focus on the issue of memory retrieval in dynamic scenes. Optimizing retrieval triggers for dynamic scenes using more accurate geometric or external supervision represents a viable path forward. We encourage the community to build upon this work to explore more comprehensive internal memory systems suitable for video world models.

\section{AI Assistance}

We used AI to polish the grammar of this paper. AI was not used to assist with the core ideas or the experimental design.



\end{document}